\documentclass[journal]{IEEEtran}

\usepackage{times,latexsym}
\usepackage{url}
\usepackage[T1]{fontenc}

\usepackage{subcaption}
\usepackage{graphicx}
\usepackage{booktabs}
\usepackage{amsthm, amsmath, amssymb}
\usepackage{graphicx, multicol} 
\usepackage{xcolor} 
\usepackage{lipsum}
\usepackage{acronym}
\usepackage[inline]{enumitem}
\usepackage{multicol} 
\usepackage{multirow}
\usepackage[numbers]{natbib}

\AtBeginDocument{%
  \providecommand\BibTeX{{%
    \normalfont B\kern-0.5em{\scshape i\kern-0.25em b}\kern-0.8em\TeX}}}

\newcommand{\OurModel}{P2-Net}

\newcommand{\circled}[1]{\raisebox{.5pt}{\textcircled{\raisebox{-.9pt} {#1}}}}

\acrodef{NLG}{Natural Language Generation}
\acrodef{TDS}{Task-oriented Dialogue System}
\acrodef{DRG}{Dialogue Response Generation}
\acrodef{CBOW}{Continuous Bag-of-Words}
\acrodef{VAE}{Variational Autoencoder}
\acrodef{NLU}{Natural Language Understanding}
\acrodef{DST}{Dialogue State Tracking}
\acrodef{PL}{Policy
Learning}
\acrodef{LSTM}{Long Short-Term Memory}

\allowdisplaybreaks

\author{
Phillip Lippe,
Pengjie Ren,
Hinda Haned,
Bart Voorn,
and~Maarten de Rijke
\thanks{Phillip Lippe, Pengjie Ren are with the University of Amsterdam, The Netherlands.
Emails: p.lippe@uva.nl, p.ren@uva.nl
}
\thanks{Bart Voorn is with Ahold Delhaize, The Netherlands.
Email: Bart.Voorn@aholddelhaize.com}
\thanks{Hinda Haned and Maarten de Rijke are with the University of Amsterdam and Ahold Delhaize, The Netherlands.
Emails: Hinda.Haned@aholddelhaize.com, m.derijke@uva.nl
}
\thanks{Pengjie Ren is the corresponding author.}
\thanks{Manuscript received August 15, 2020; revised XX, XX.}}

\begin{document}

\title{Diversifying Task-oriented Dialogue Response Generation with Prototype Guided Paraphrasing}

\markboth{IEEE/ACM Transactions on Audio, Speech, and Language Processing}%
{Lippe \MakeLowercase{\textit{et al.}}: Diversifying Task-oriented Dialogue Response Generation}

\maketitle

\begin{abstract}
Access to information is increasingly conversational in nature.
\acp{TDS} aim to help users achieve a specific task through conversations, e.g., booking a flight.
\ac{DRG} is one of the core \ac{TDS} components; it translates system actions (e.g., \texttt{request(phone)}) into natural language responses (e.g., Can I have your phone number, please?).
  
Existing methods for \ac{DRG} in \acp{TDS} can be grouped into two categories: template-based and corpus-based.
The former prepare a collection of response templates in advance and fill the slots with system actions to produce system responses at run-time.
The latter generate system responses token by token by taking system actions into account.
While template-based \ac{DRG} provides high precision and highly predictable responses, they usually lack in terms of generating diverse and natural responses when compared to (neural) corpus-based approaches.
Conversely, while corpus-based \ac{DRG} methods are able to generate natural responses, we cannot guarantee their precision or predictability.
Moreover, the diversity of responses produced by today's corpus-based \ac{DRG} methods is still limited.
  
We propose to combine the merits of template-based and corpus-based \acp{DRG} by introducing a \textbf{p}rototype-based, \textbf{p}araphrasing neural \textbf{net}work, called \OurModel{}, which aims to enhance quality of the responses in terms of both precision and diversity.
Instead of generating a response from scratch, \OurModel{} generates system responses by paraphrasing template-based responses.
To guarantee the precision of responses, \OurModel{} learns to separate a response into its semantics, context influence, and paraphrasing noise, and to keep the semantics unchanged during paraphrasing.
To introduce diversity, \OurModel{} randomly samples previous conversational utterances as prototypes, from which the model can then extract speaking style information.
We conduct extensive experiments on the MultiWOZ dataset with both automatic and human evaluations.
The results show that \OurModel{} achieves a significant improvement in diversity while preserving the semantics of responses.
\end{abstract}


\IEEEpeerreviewmaketitle


\section{Introduction}

\IEEEPARstart{T}{ask-oriented} dialogue systems (TDSs\acused{TDS}) are becoming more widespread~\cite{10.1145/3184558.3191539,10.1145/3317612,10.5555/2817174.2817185}; they are being used in a wide range of applications, including information seeking, shopping assistants and chat bots~\cite{10.1145/3209978.3210183,shum2018eliza}.
A typical \ac{TDS} system usually includes a pipeline of several modules, e.g., \ac{NLU}, \ac{DST}, \ac{PL}, and \acf{DRG}~\cite{10.1109/TASLP.2019.2919872,10.1109/TASLP.2019.2949687}.
\ac{DRG} is one of the core \ac{TDS} components; it translates system actions (e.g., \texttt{request(phone)}) into natural language responses (e.g., Can I have your phone number, please?).
Existing methods for \ac{DRG} in \acp{TDS} can be grouped into two categories: \emph{template-based} (or rule-based) and \emph{corpus-based}~\cite{DialogueSystemSurvey}. 
Template-based approaches use a set of manually created response templates, which are instantiated with slot values at run-time. 
Such approaches tend to produce high-precision results with a high degree of predictability. 
A drawback of these approaches is a low degree of diversity, which may result in unnatural conversations due to the use of a fixed and limited set of response templates. 
In contrast, corpus-based approaches can directly generate responses token by token at run-time and thereby generate responses that tend to be diverse and fluent. 
However, corpus-based approaches may generate unexpected responses due to the complexity of the \ac{TDS} task and the unpredictable nature of current (imperfect) \ac{NLG} techniques.

How can we combine the strengths of template-based and corpus-based approaches to response generation?
We propose a method to refine responses produced by a template-based system with a corpus-based model based on a combination of neural prototype editing~\cite{PrototypeEditing2} and paraphrasing techniques~\cite{lan-xu-2018-neural}. 
We assume that a response is generated to consider three components: the \emph{semantics} (i.e., what to say), the \emph{context style} (i.e., the user's question and previous dialogue turns), and \emph{paraphrasing noise} (i.e., unnecessary words, rephrasing). 
The semantics can be determined best by a template-based approach, as for \acp{TDS} there is usually a limited set of tasks. 
In contrast, the context style and paraphrasing noise must be flexible as there are many ways of expressing the same meaning, e.g., using different sentence functions. 
Hence, a corpus-based approach is best suited for these components. 
By rephrasing the template-based response with a corpus-based model, we want to keep the high controllability and precision of a typical template-based approach, while generating more diverse responses and natural conversations as in corpus-based approaches to response generation. 

\begin{figure}
    \centering
    \includegraphics[width=\columnwidth]{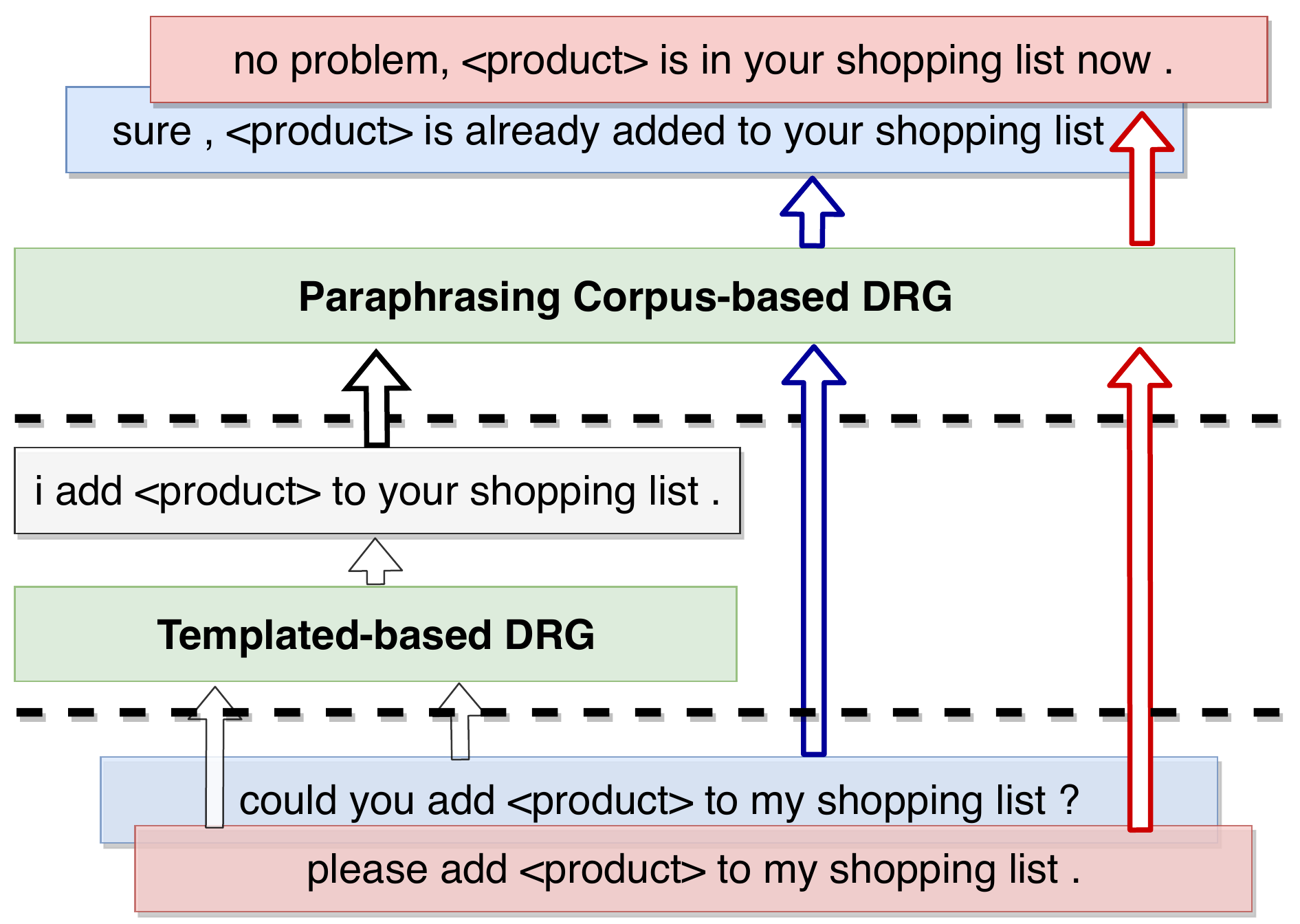}
    \caption{Overview of the combined template-based and corpus-based approach to response generation. 
    First, a template-based dialogue system generates a template response based on the user's question. 
    In a second step, the response is refined by a paraphrasing model that takes the conversational context into account.}
    \label{fig:introduction_task_system_overview}
\end{figure}

Figure~\ref{fig:introduction_task_system_overview} illustrates the combined strategy that we propose.
We propose to take both the context and a template response as input, and generate a new, context-aware response. 
Specifically, given a user query and its corresponding template response (from a template-based \ac{TDS} system), our goal is to paraphrase the template response to increase its diversity. 
This task is significantly simpler than generating a response from scratch, thereby allowing us to focus on style details.
This task differs from previous work on diversifying text generation through style transfer~\cite{StyleTransferNonParallel, StyleTransferUnsup}, which aims to rewrite a sentence with a target style, while keeping the semantics mostly unchanged. 
In our task, it is not sufficient to simply adjust to the style of the user because we need to establish a natural conversation with filling words like ``sure'' or ``of course.'' 
It also differs from traditional paraphrasing~\cite{ParaphraseDPage, ParaphraseCounterfactualDebiasing} as we should not just diversify the templates, but also incorporate the conversational context. 

In this work, we propose a \textbf{p}rototype-based, \textbf{p}araphrasing neural \textbf{net}work, called \OurModel{}, to achieve the process in Figure~\ref{fig:introduction_task_system_overview}.
The \OurModel{} learns to encode the three response components independently, i.e. semantics, context style and paraphrasing noise. 
%
Specifically, given a conversational context and a template response, the \OurModel{} first encodes the template response into a \textit{response semantic vector}.
Then, it samples a set of \textit{context prototypes} and encodes them into a \textit{context style vector} guided by the context.
Similarly, it samples a set of \textit{response prototypes} and encodes them into a \textit{response paraphrasing noise vector} guided by the ground truth response during training and randomly sampled during evaluation.
The prototypes are jointly learned and we expect each prototype represents a particular latent style or noise. 
We strongly limit the information flow from the ground truth response, ensuring that the ground truth response can only help extract the latent style information (namely the paraphrasing noise) from response style prototypes that it cannot retrieve from the other sources. 
As the ground truth response can only determine what style information from the prototypes is used, it is only used during training and during inference, we sample from the prototypes instead.
To ensure the semantic preservation from the template response, we also introduce an inductive bias on using all the provided slots.
Using a set of slots as input, we remove a slot from the input once the network has selected it as the next word prediction. 
As in the training set, the network has only seen sentences stop when the set of slots is empty, it will inherently transfer this knowledge to the generation of new responses.
\OurModel{} is trained on the task of generating the ground truth responses.
%
As no sufficiently large dataset of aligned template-based and corpus-based responses exists, we propose a weakly-supervised learning mechanism to train \OurModel{}, where we assume system responses with the same system actions are paraphrases, which have the same semantic while different styles.
We show in experiments that \OurModel{} is able to split the semantic and the style of a response into separate parts, and keep the high semantic precision of the template-based approach, while generating more diverse responses by varying response style prototypes.
We compare \OurModel{} to stochastic beam search (an effective method to promote diverse responses), and find that \OurModel{} can outperform stochastic beam search by a large margin in terms of diversity.
We also conduct human evaluation to confirm that \OurModel{} achieve much better diversity performance without hurting the quality of generated responses, i.e., grammaticality, naturalness, semantics, as well as context awareness.

In summary, the contributions of this paper are:
\begin{itemize}[leftmargin=*,nosep]
    \item We propose a new workflow for \acf{DRG} in \acf{TDS} by combing template-based and corpus-based \ac{DRG} methods.
    \item We propose \OurModel{} with neural prototype guided paraphrasing to achieve the workflow, which is one of the first proposals to use prototype editing for style adjustment in the context of \ac{TDS}. 
    \item We devise a weakly-supervised learning mechanism for splitting the semantics and the style of a response into separate parts. We also introduce an inductive bias on using all the provided slots, which ensures a success rate of 100\%.
    \item We conduct both automatic and human evaluations to show the effectiveness of \OurModel{} in terms of the diversity and quality of generated responses.
\end{itemize}


\section{Related Work}
\label{sec:related_work}

\subsection{Dialogue response diversity}
Diversifying the responses produced by conversational agents is a topic of growing interest~\cite{DiversityPromotingObjective, DiversityPromotingGAN, DiversityHighQualitySeq2Seq, DiversityFrequencyAwareCE, DiversityAdversarialInformationMaximization}. 
There have been many approaches to reach this goal. 
One is to adjust the loss function or learning mechanism to encourage diversity. 
\citet{DiversityPromotingObjective} argue that the standard maximum likelihood training might favor frequent responses, and hence leads to low diversity. 
Instead, they propose to use the maximum mutual information which promotes outputs that are specific for a certain input. 
Nevertheless, this method requires an inverse model being trained by swapping the inputs and outputs, or relies on beam search. 
\citet{DiversityFrequencyAwareCE} suggest to weight tokens in the cross-entropy loss based on their frequency such that rare words will be weighted higher. 
While this method increases token diversity, it might promote other diversity aspects like sentence structure or phrasal paraphrasing.
Some studies adopt generative adversarial networks~\cite{GAN}, where the discriminator is used to distinguish between real and fake samples. 
For the text domain, this min-max game is usually reformulated as a reinforcement learning objective with the discriminator's output being the reward of the generator~\cite{DiversityPromotingGAN, DiversityAdversarialLearning}. 
While this approach has been shown to generate more human-like responses, training can be very unstable and may not boost the results as much as expected~\cite{DiversityAdversarialLearning}.
Besides, the methods listed above have all (initially) been proposed for chitchat and cannot be applied to \ac{TDS} directly, as they cannot guarantee to preserve the semantics of the responses, which is fine in chitchat scenarios but not acceptable in \acp{TDS}~\cite{ren-2020-thinking}.

A different approach to diversifying responses is to change the way one samples each token from each decoding step.
A commonly used method is beam search~\cite{BeamSearch}, where the responses are generated by following the $B$ most probable token-sequences so far. 
To enable greater diversity, \citet{DiversityHighQualitySeq2Seq} propose a stochastic variant of beam search; instead of taking the top $B$ beams, one samples tokens based on the output probability and samples beams accordingly. 
\citet{DiverseBeamSearch} split up the beams into $G$ groups, and add a penalty term for selecting tokens or n-grams that occur in other beam groups. 
The benefit of these methods is that they can be applied to any, already trained sequence-to-sequence generation models.

\subsection{Paraphrasing}
Paraphrasing refers to the task of detecting and generating paraphrases.
In this work, we survey paraphrase generation.
Conventional approaches model paraphrase generation as a supervised encoding-decoding process~\cite{prakash-etal-2016-neural}. 
For example, given a sentence, \citet{DBLP:conf/aaai/GuptaASR18} propose to combine a \ac{VAE} with a \ac{LSTM} to generate paraphrases.
\citet{li-etal-2018-paraphrase} present a deep reinforcement learning approach to paraphrase generation. 
Their model consists of a generator and an evaluator. 
The generator produces paraphrases given a sentence. 
The evaluator judges whether two sentences are paraphrases of each other. 
The generator is first trained by supervised learning and then fine-tuned by reinforcement learning in which the reward is given by the evaluator.
\citet{qian-etal-2019-exploring} also adopt reinforcement learning to paraphrasing to promote the diversity of generated paraphrases, where they design two discriminators and multiple generators to generate a variety of different paraphrases.
On the other hand, some studies investigate weakly-supervised paraphrasing by synthesizing pseudo-paraphrase pairs~\cite{wieting-gimpel-2018-paranmt}.
For example, \citet{lan-etal-2017-continuously} present a method to collect paraphrases from Twitter by linking tweets through shared URLs.
There are also some unsupervised paraphrasing studies~\cite{barzilay-lee-2003-learning}.
For example, \citet{liu-etal-2020-unsupervised} model paraphrase generation as an optimization problem and consider semantic similarity, expression diversity, and language fluency to define the learning objective.
Paraphrasing has also been applied to various tasks to boost performance, such as machine translation~\cite{10.3115/1219840.1219914}, information retrieval~\cite{10.3115/1072228.1072389} as well as dialogue systems~\cite{shah-etal-2018-bootstrapping}.
For example, \citet{gao-etal-2020-paraphrase} propose a framework that jointly learns a paraphrase model and a response generation model to improve the dialog generation performance.

What we add on top of the related work discussed above is that we propose a new schema to diversify \ac{DRG} in \acp{TDS} based on prototype editing and paraphrasing.
The idea of prototype editing is to first sample a prototype sentence from the training corpus and then edit it into a new sentence, instead of generating a sentence from scratch~\cite{PrototypeEditing2}.
The prototype sentences have different styles so that we expect to get diverse responses w.r.t. different prototype sentences.
The idea of paraphrasing is to rephrase a sentence in different styles without changing its semantics~\cite{ParaphraseDPage,qian-etal-2019-exploring}.
We use paraphrasing to make sure that the semantics of the rephrased is kept unchanged.
There have been some applications of prototype editing and paraphrasing (separately) in information retrieval~\cite{10.1145/2911451.2914800,NIPS2018_8209,10.1145/3357384.3358102} and natural language processing tasks~\cite{ParaphraseCounterfactualDebiasing,PrototypeEditing,li-etal-2019-decomposable}.
But to the best of our knowledge, none of them has proposed to combine prototypee editing and paraphrasing to diversify \ac{DRG} in \acp{TDS}.


\section{Method}
\label{sec:model}

\subsection{Task formulation}

Given a template response (from a template-based \ac{TDS} system) and a dialogue context (from previous turns), the task is to paraphrase the template response to (1) keep its semantics unchanged, and (2) increase its diversity. 
In order to keep the semantics unchanged, we need to make sure all slots of the template response are covered and placed in the right position of the response.
In order to increase the diversity, we need to make the response aware of context and incorporate random noise that can only influence the non-essential content of the response.


\subsection{Overview of \OurModel{}}
\label{sec:model_overview}
 
A human-like response of a \ac{TDS} is influenced by various factors, which we can group in three main components, namely the \textit{semantics}, \textit{context style}, and \textit{paraphrasing noise}. 
The \textit{semantics} of a response determines the content or message to communicate to the user, and template-based \acp{TDS} perform especially well on it. 
However, there are various ways to express the same semantics. 
On the one hand, it is influenced by the \textit{context style}, i.e., the preceding conversation and the question of the user. 
Depending on the specific way the user is asking his or her questioon, we can respond more naturally. 
For example, if the question is ``Can you tell me the name of the hotel?'', the \ac{TDS} could respond with ``I absolutely can, the name is \ldots'' while this starting phrase is not suitable for all questions.
Even if the context turns some of the paraphrases inappropriate, there are still some possible sentence variations, which we summarize by the term \textit{paraphrasing noise}, i.e., unnecessary words, e.g., ``sure'', ``of course.''

\begin{figure*}
    \centering
    \includegraphics[width=\textwidth]{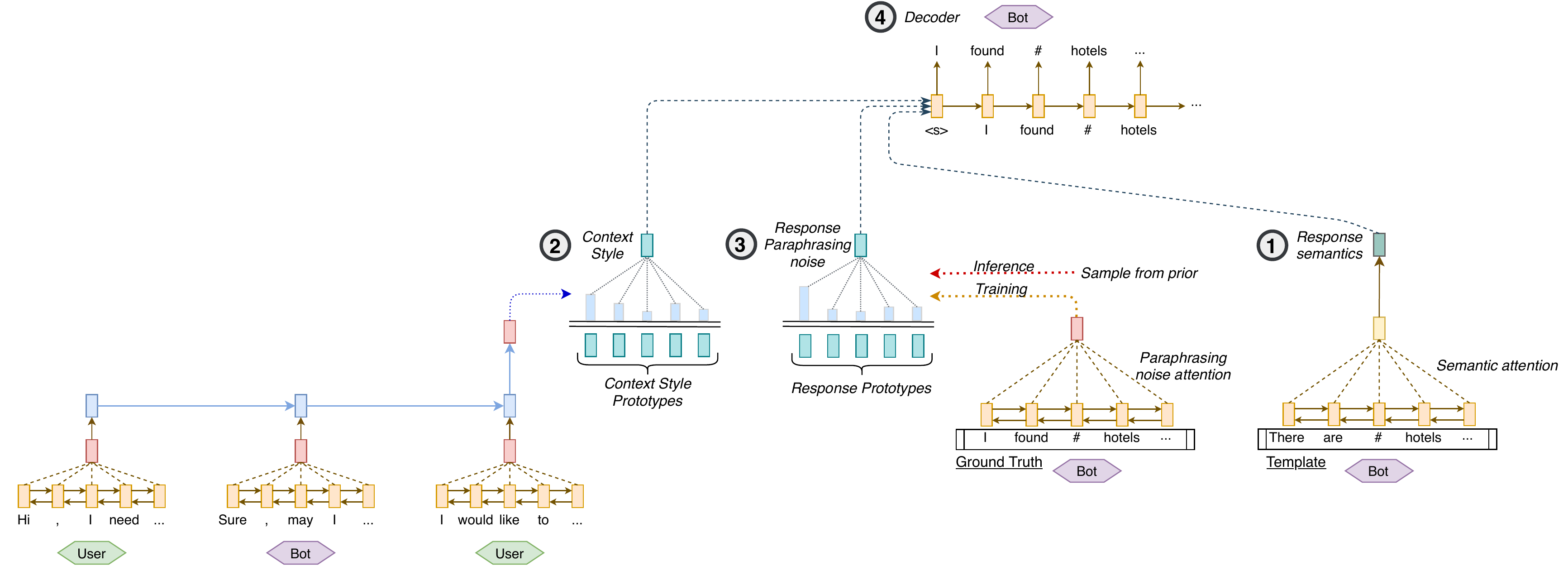}
    \caption{
    	Visualization of the response generation process within \OurModel. \S\ref{sec:model} contains a walkthrough of the model.}
    \label{fig:model_prototype_encoder_overview}
\end{figure*}

Based on this understanding of response generation, and to model each of these three components independently, we propose our context-aware paraphrasing model, \OurModel, which is sketched in Figure~\ref{fig:model_prototype_encoder_overview}. 
The input template response is encoded by a Bi-LSTM into a \textit{response semantic vector} (\circled{1} in Figure~\ref{fig:model_prototype_encoder_overview}) constituting a feature vector. 
The \textit{context style vector} (\circled{2}) and \textit{paraphrasing noise vector} (\circled{3}) are represented by modeling \textit{context prototypes} and \textit{response prototypes} from which the model can select a weighted sum. 
All three vectors are input to the decoder (\circled{4}). 
The goal is to generate diverse responses while being able to alternate the style without changing the semantics. 
To learn the split between semantics, context style and paraphrasing noise, the model is trained to predict the next response in a conversation given different inputs for each of the components.
The semantics of the response is modeled by encoding the output of a template-based \ac{TDS} for the corresponding conversation (\circled{1}). 
The context component is extracted from previous conversation turns by the user and the \ac{TDS} (\circled{2}). 
Although models that purely rely on the dialogue context can perform reasonably well in response generation \cite{pei-2020-ecai}, we expect the model to rely on the template for the semantics as it is a much easier task. 
Paraphrasing noise cannot easily be predicted on external inputs as it is based on random choice. 
We therefore propose to model it from the ground truth directly while limiting the information flow to prevent \OurModel{} from simply copying the response (\circled{3}).

During training, \OurModel{} learns to generate responses based on a template, a dialogue context and the ground truth. 
After training, we replace the ground truth with a sampling mechanism to obtain paraphrasing noise inputs. 
This setup is expected to generate more diverse outputs than post-processing methods such as beam search because the model explicitly learns different styles of paraphrasing.
	    
Next, we give a detailed explanation of each part, i.e., embeddings in \ref{sec:model_slot_embeddings}, the encoder in \ref{sec:model_encoder}, the decoder in \ref{sec:model_decoder}, and the learning process in \ref{sec:model_learning}.

\subsection{Embeddings}
\label{sec:model_slot_embeddings}
There are two types of embedding: word embeddings and slot embeddings.
For the word embeddings, we use GloVe \cite{pennington2014glove} as initialization and fine-tune them during training.

\begin{figure}[t!]
    \centering
    \includegraphics[width=0.4\textwidth]{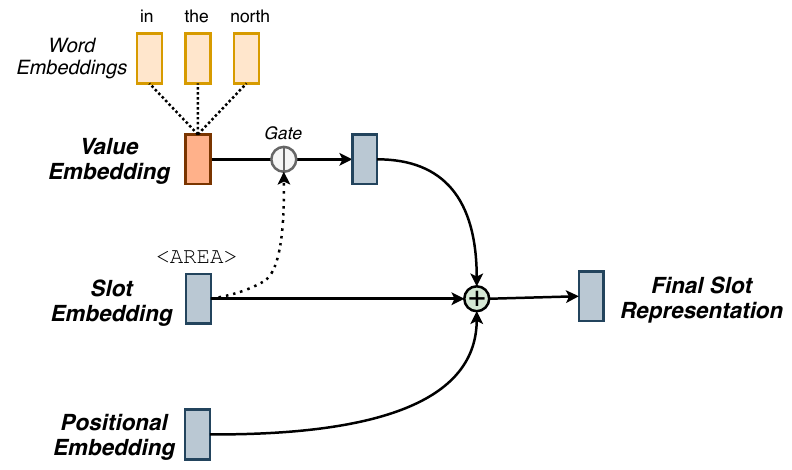}
    \caption{Embedding of a template slot. Each slot is represented based on an embedding of its type (e.g., area, name), its position, and its value (e.g., the actual words in the response). A gate is applied on the value embedding based on the slot type to filter out unnecessary information.}
    \label{fig:model_slot_embeddings}
\end{figure}

A template from the template-based \ac{TDS} provides slots in which specific information such as restaurant names or phone numbers are stored. 
Paraphrasing a template requires an understanding of these slots, and hence they should be taken differently as word embeddings and need to be properly embedded in the neural model.
To represent the given slots in a template, three components are necessary. 
The approach is visualized in Figure~\ref{fig:model_slot_embeddings}.

First, to recognize the general semantics of a slot, we learn an embedding for each type (e.g., area, name, etc.), similarly to any word embedding. 
Second, we also need to distinguish between slots with the same type in case we have a template with, for example, multiple restaurant names. 
Furthermore, the order of slots can be important as well: if we have two names and two addresses, the network needs to reason about which name belongs to which address. 
To implement this ordering, we use a sinusoidal position embedding~\cite{VaswaniAttention}. 
Note that the position is counted only for slots with the same type, and not determined by the overall order of all slots. 
Learned embeddings suffer from the issue that cases of more than 3 slots of the same type are rare, and hence harder to learn. 
Third, the actual value of the slot is relevant as well to form a natural sentence. 
For example, ``the'' should not be used in front of a name if the name already starts with ``the''. 
We choose a rather simple approach to embed the values, namely a single-layer \ac{CBOW} with a gate modeled by the slot type embedding. 
While the \ac{CBOW} prevents any strong overfitting on the slot values, the gate controls how much information is necessary to improve the slot representation. For instance, a phone number does not need any value embedding as it usually does not influence the sentence structure.

All three components added together result in the final representation that is used in both the encoder and decoder.

\subsection{Encoder}
\label{sec:model_encoder}

The encoder aims to generate the semantic and style representations for the desired output response.
For each representation, we specify a certain architecture, which we will discuss next.

\subsubsection{Semantic encoding}
The template response (\circled{1}) is used to encode the semantics by using a one-layer Bi-LSTM \cite{LSTM} network with global attention, using the last hidden state $h_{\text{end}}$. 
The attention can be specified as follows:
\begin{equation}
\label{semantic_att}
\begin{split}
s_{\text{semantics}} & = \frac{\sum_{t=1}^{T} h_t \cdot \exp\left(\text{attn}(h_t;h_{\text{end}})\right)}{\sum_{t=1}^{T} \exp\left(\text{attn}(h_t;h_{\text{end}})\right)} \\
\text{attn}(h^{(i)};h^{(T)}) & = \tanh(W_h h^{(i)} + W_c h^{(T)} + b_{\text{attn}}),
\end{split}
\end{equation}
where $h_t$ is the hidden state of the Bi-LSTM at timestep $t$. 
We refer to the output feature vector, $s_{\text{semantics}}$, as \textit{response semantic vector}, and the attention distribution as \textit{semantic attention} of a response.

\subsubsection{Context style encoding}
The \textit{context style} is encoded with a hierarchical RNN on a limited number of previous conversation turns (\circled{2}). 
We use the same one-layer Bi-LSTM network as for semantic encoding, but with an attention module with separate weights, which we refer to as \textit{context style attention}.
We could use the output of the LSTM directly as a representation for the context as most existing works do~\cite{10.1145/3317612}.
However, the context is high-dimensional and expected to be very noisy as a lot of information is not needed in order to diversify the next response.
For example, the semantics can be obtained from the template response, so to a large extent, it is not necessary to encode context semantics.
To this end, we devise a prototype layer.
We do this by introducing a fixed set of learnable embeddings, which we call \emph{context prototypes} $p^c_1, \ldots, p^c_K$ (\circled{2}). 
The context style vector is a weighted sum of these prototypes. 
The weights are determined by the context using an attention module:
\begin{equation}
\label{context_att}
\begin{split}
\hat{s}^{\text{context}}_{\text{style}} = \frac{\sum_k p^c_k \cdot \exp(\text{attn}(p^c_k;s_{\text{context}}))}{\sum_k \exp(\text{attn}(p^c_k;s_{\text{context}}))},
\end{split}
\end{equation}
where $p^c_k$ are the context prototype vectors and $s_{\text{context}}$ is the encoded feature vector of the context (the last hidden state of LSTM).
The prototype layer prevents the model from encoding a significant amount of unnecessary information into the feature vector, and thus might help to generalize better. 
We show the merit of this design choice in \ref{s05-2}.
We found that taking more than three previous turns into account, as visualized in the bottom left hand side corner of Figure~\ref{fig:model_prototype_encoder_overview}, did not further improve the performance.

\subsubsection{Paraphrasing noise encoding}
As motivated previously, paraphrasing noise can only be determined by the ground truth response. 
Hence, during training, we encode the ground truth into a feature vector representing the \textit{response paraphrasing noise} (\circled{3}). 
We use the same one-layer Bi-LSTM as for the template encoding, but with an attention module with separate weights, which we refer to as \textit{paraphrasing noise attention}.
However, if we would directly use the attention output as input to the decoder for generation, the network would attempt to encode the whole ground truth response, and completely ignore the semantics from the template response. 
Hence, we need to create a bottleneck to limit the information flow from the ground truth. 
Again, we do this by introducing a fixed set of \emph{response prototypes} $p^r_1, \ldots, p^r_K$ (\circled{3}). 
Instead of allowing an arbitrary representation, the response paraphrasing noise vector is a weighted sum of these prototypes. 
The weights are determined by the ground truth response using another attention module:
\begin{equation}
\begin{split}
\hat{s}^{\text{response}}_{\text{noise}} = \frac{\sum_k p^r_k \cdot \exp(\text{attn}(p^r_k;s_{\text{response}}))}{\sum_k \exp(\text{attn}(p^r_k;s_{\text{response}}))},
\end{split}
\end{equation}
where $p^r_k$ are the prototype vectors and $s_{\text{response}}$ is the encoded feature vector of the ground truth (the last hidden state of LSTM). 
Thus, the ground truth can guide the generation process by providing information that cannot be extracted from the template and context, namely the paraphrasing noise, but not enough for the generation process to fully reconstruct the response solely from this representation. 

During evaluation, the ground truth response is not available. 
To obtain diverse responses, we can simply sample from the \textit{paraphrasing noise attention}, for instance with a Dirichlet distribution. 
We expect that different sampling results (combinations of response prototypes) will lead to different responses with the same semantics, but different ways of expressing it.
 
\subsection{Decoder}
\label{sec:model_decoder}

\begin{figure}[t]
    \centering
    \includegraphics[width=\columnwidth]{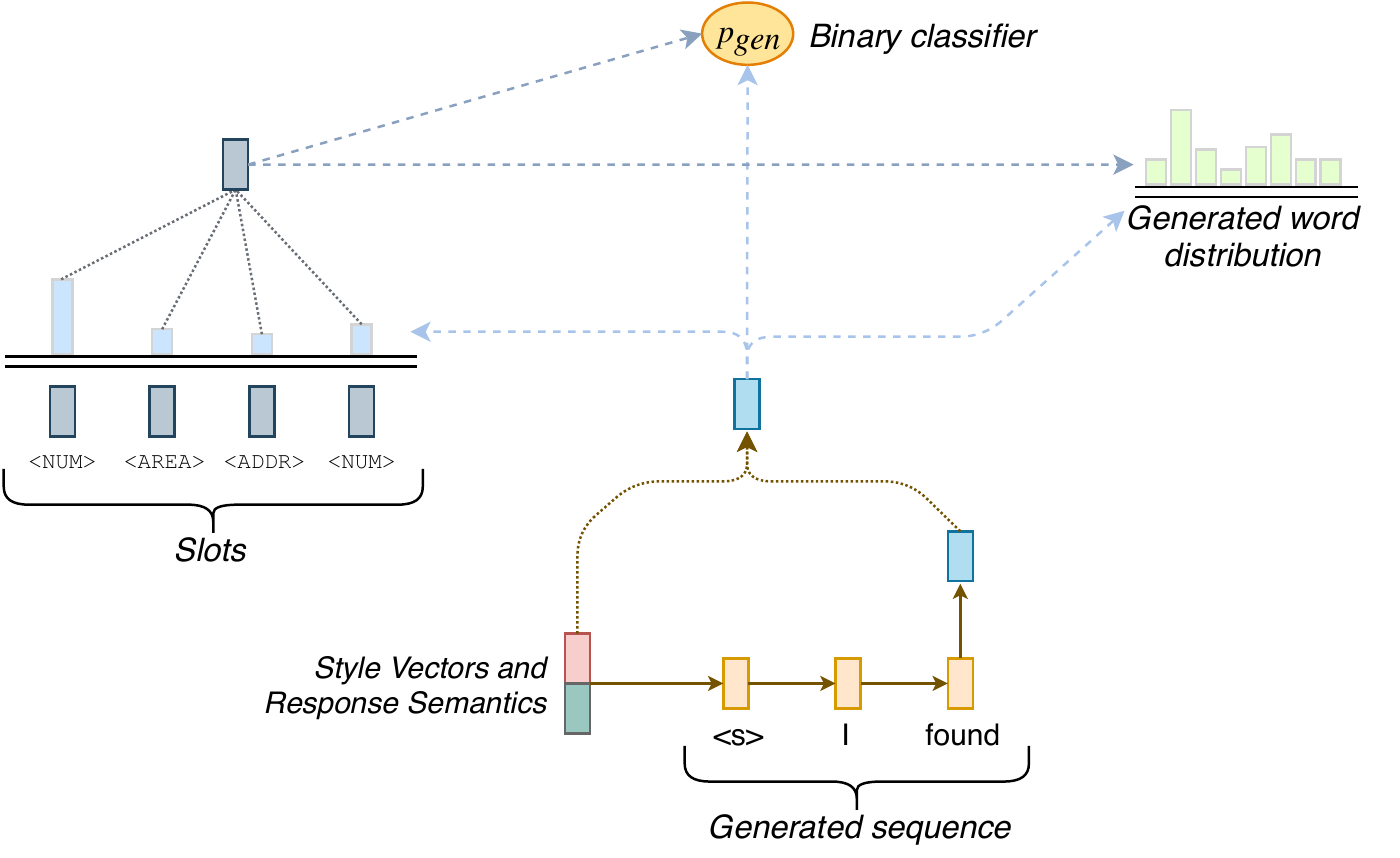}
    \caption{At each generation step, the decoder determines an attention distribution $p_{\text{slot}}$ over slots based on the current hidden state $h_t$ and the semantic and style vectors. This is being used to predict the probability of generating a new word, $p_{\text{gen}}$, and the corresponding word distribution $p_{\text{word}}$.}
    \label{fig:model_decoder_overview}
\end{figure}

Based on the semantics $s_{\text{semantics}}$, the context style vector $\hat{s}^{\text{context}}_{\text{style}}$ and the response paraphrasing noise vector $\hat{s}^{\text{response}}_{\text{noise}}$, the decoder generates a new response specific to the inputs. 
The module is inspired by the pointer network architecture \cite{PointerNetwork, PointerNetworkSummarization}, and consists of a one-layer unidirectional LSTM as base network. 
See Figure~\ref{fig:model_decoder_overview} for an overview of the decoder architecture.

The initial state is generated based on the encoded context style and semantics.
Similar to the pointer network, we use the current state $h_t$ as context vector to determine an attention distribution $p_{\text{slot}}$ over the slots that should be included in the output response. 
The weighted sum of the slot embeddings (see \S\ref{sec:model_slot_embeddings} for details) is used as an additional input for determining the output distribution $p_{\text{word}}$ over words. 
Furthermore, a binary classifier is applied to determine whether the next word should be generated from the vocabulary ($p_{\text{gen}}=1$), or a slot should be used instead ($p_{\text{gen}}=0$). 
The probability is calculated as follows:
\begin{equation}
\begin{split}
\mbox{}\hspace*{-1mm}
&p_{\text{gen}} = {}\\
\mbox{}\hspace*{-1mm}
&\sigma\!\left(w_h h^{D}_t + w_{\text{s}} \hat{s}_{\text{semantics}} + w_{\text{c}} \hat{s}^{\text{context}}_{\text{style}} + w_{\text{gt}} \hat{s}^{\text{response}}_{\text{noise}} + b_{\text{gen}}\right)\!,
\end{split}
\end{equation}
where $h^{D}_t$ is the hidden state of the decoder at timestep $t$; and $\hat{s}_{\text{semantics}}$, $\hat{s}^{\text{context}}_{\text{style}}$ and $\hat{s}^{\text{response}}_{\text{noise}}$ are the encoded response semantics, context style vector and the paraphrasing noise vector, respectively.

During inference and sampling, we experienced that obtaining a probability distribution over all tokens, i.e., multiplying $p_{\text{gen}}$ with the probabilities over the vocabulary and $1-p_{\text{gen}}$ with the attention distribution over the slots, strongly favors the slots. 
To counteract this behavior, we generate a new word if $p_{\text{gen}}>\delta$ where $\delta$ represents a threshold, and otherwise select a slot based on its corresponding distribution.
A grid search over different thresholds has shown that $\delta=0.5$, i.e., the center point of the scale, leads to stable and good results.

Another important aspect of the slots is that in most responses, each slot is only used once in a prediction. 
In our dataset (see \ref{sec:experiments_dataset}), we experienced that almost 99\% of the answers given by a human contained each slot only once. 
Therefore, we expect the network to learn using each slot once as well.
However, it might be hard for the decoder to remember whether it has already used a certain slot or not, which may lead to repetitive outputs. 
To prevent this, we introduce an inductive bias by masking out the slots that have already been used in the output. 
During training, we mask the slots based on the ground truth, while for inference, we do it when the network predicts a slot. 
With this technique, the task of integrating the slots is simplified and simpler patterns can be learned, such as that a sentence can only end once all slots have been used in the prediction. 

\subsection{Learning}
\label{sec:model_learning}

Given a conversation context, a template, and a ground truth response, we train \OurModel{} to reconstruct the ground truth response. 
We consider responses with the same dialogue action and the same slots (types and amount, not actual values) as paraphrases in different contexts.
So for a given ground truth response, its paraphrases are considered as templates.
Specifically, let $y^{(i)}$ denote whether the token at position $i$ of the ground truth response is a slot ($y^{(i)}=0$) or a word ($y^{(i)}=1$).
Then, the loss for binary classifier $p_{\text{gen}}$ can be defined as:
\begin{equation}
\mbox{}\hspace*{-1mm}
\mathcal{L}_{\text{gen}} = -\sum_{i} y^{(i)}
\left(
\log p^{(i)}_{\text{gen}} + \left(1 - y^{(i)}\right)\log \left(1 - p^{(i)}_{\text{gen}}\right)
\right)\!.
\hspace*{-1mm}\mbox{}
\end{equation}
If $y^{(i)}=0$, i.e., the token is a slot, we add the negative log likelihood of that slot in the decoder's attention distribution $p_{\text{slot}}$.
In case $y^{(i)}=1$, i.e., the token is a word, we add the negative log likelihood of the word in the decoder's output distribution $p_{\text{word}}$. 
\begin{equation}
\begin{split}
&\mathcal{L}^{(i)}_{\text{word}}=
\begin{cases}
-\log p^{(i)}_{\text{slot}} & \text{if }{y^{(i)}=0} \\
-\log p^{(i)}_{\text{word}} & \text{if }{y^{(i)}=1}. \\
\end{cases}
\end{split}
\end{equation}
The final loss is a combination of $\mathcal{L}_{\text{gen}}$ and $\mathcal{L}^{(i)}_{\text{word}}$:
\begin{equation}
\mathcal{L}_{\text{final}} = \mathcal{L}_{\text{gen}} + \sum_{i} \mathcal{L}^{(i)}_{\text{word}}.
\end{equation}


\section{Experimental Setup}
\label{sec:experiments}

We seek to answer the following research questions:
\begin{enumerate}[label=(RQ\arabic*), leftmargin=*, nosep]
	\item Can \OurModel{} generate more diverse responses than post-processing methods? And which variant of \OurModel{} performs best?
	\item Is \OurModel{} able to paraphrase a template without changing its semantics?
	\item Can \OurModel{} learn to attend to tokens w.r.t. semantics with semantic attention and tokens w.r.t. speaking styles with context style attention?  
	\item How diverse are the responses of \OurModel{} demonstrated with qualitative analysis? What are the typical failures?
\end{enumerate}

\subsection{Dataset}
\label{sec:experiments_dataset}
Based on the sketched use-case of our model in Figure~\ref{fig:introduction_task_system_overview}, an ideal dataset would contain conversations of a human to a template-based \ac{TDS}, with paraphrased responses by another human that take the previous conversation into account and do not sound like they were automatically generated.
However, creating such a dataset is expensive and is not guaranteed to provide diverse, natural conversations as the human user might have known that he/she was interacting with a dialogue system. 

Thus, to ensure that we train on human responses that fit the context and have natural conversations, we require dialogues between two humans where one replaces the automated dialogue system.
We therefore perform our experiments on the MultiWOZ dialogue dataset~\cite{MULTIWOZ}, which contains human-to-human conversations across multiple domains. 
Every response is annotated with a dialogue action and the slot entities (e.g., the name of a hotel) used in the sentence. 
To obtain our templates, we group responses with the same dialogue action and the same slots (types and amount, not actual values) as paraphrases in different contexts. 
In this set, we can use any sentence to represent the template for another sentence because they are expected to have the same semantics. 
If a response has more than one sentence and/or dialogue actions, we split it to prevent the mixture of multiple semantics. 
Furthermore, to counteract overfitting, we only consider response sets with at least 4 responses, as otherwise the network can learn an almost 1-to-1 mapping between template and output. 
Overall, this leaves us with 1,147 sets of different dialogue actions and/or slots, and about 68,000 responses.

Note that certain sets contain many more responses than others, as, e.g., the dialogue action ``general request more'' has over 12,000 instances. 
To prevent the model from focusing only on those responses, we balance the training set by controlling the frequency with which examples from a dialogue action are shown. 
A suitable scheduling has been found to take a frequency proportional to the square root of the number of instances for a dialogue action, with an upper limit of 200. 
Hence, the frequency of the dialogue action ``general request more'' is reduced from 17.6\% to about 1.8\%, allowing the model to focus on learning a variety of different response types.

The validation and test datasets are built up from 100 response sets for which the network has already seen examples (but different contexts and responses), and 100 sets with a new, unseen dialogue action. 
All sets have five to seven responses. 
Hence, we test whether the network can generalize to new contexts and to new dialogue actions/template semantics. 

\subsection{Baselines}
As a baseline, we perform beam search on the output with a beam size of $N$. 
Standard beam search has been shown to give less diverse results~\cite{DiverseBeamSearch}, and various extensions like stochastic beam search \cite{DiversityHighQualitySeq2Seq} have been proposed to alleviate this issue. 
Nevertheless, we experienced that those struggled with the slot generation as a purely probabilistic interpretation of the decoder tends to favor the slots, which we will discuss in detail in \ref{s05-4}.
Especially the hyperparameters of diverse beam search, such as the penalty for using the same beam $\gamma$, were hard to optimize. 
For us, the best beam search method was stochastic beam search, possibly due to its sampling behavior, which also introduces diversity by incorporating random noise. 

To further setup a baseline, we train \OurModel{} with two configurations. 
The first configuration is \OurModel{} with context and slots as inputs, and the second is \OurModel{} with context, slots and template as inputs.
Note that, for these two configurations, we do not use the context style prototypes and  prototype layer is thereby removed and the context style prototypes, which represents standard setups for response generation on the MultiWOZ dataset, except that we provide the slots and/or templates to include in the response instead of a database~\cite{MULTIWOZ}. 

\subsection{Diversity evaluation}
\label{sec:experiments_diversity_evaluation}

A commonly used metric for diversity is the proportion of unique uni-/bigrams compared to the overall sentence lengths \cite{DiversityPromotingObjective}:
\begin{equation}
\text{Distinct-n} = \frac{\left|\bigcup_{n=1}^{N} \mathcal{W}_n\right|}{\sum_{n=1}^{N} \left|\mathcal{W}_n\right|},
\end{equation}
where $\mathcal{W}_n$ denote the set of uni- or bigrams in the sample $n$, and $|\mathcal{W}_n|$ the number of elements in this set. 
While the unigram score shows the diversity at the word level, bigrams can grasp small structural differences between the responses. 
Note that we view each slot as a single token, independent of the size of its content. 
So for example, the slot \texttt{name="West Side Hotel"} is counted as a single token and not three, as this slot is expected to occur in all responses, and hence the slots do not contribute to the diversity of responses.

\subsection{Semantic evaluation}

Besides diversity, it is also important to evaluate the coherence and textual correctness  among generated responses. 
Diversity can be maximized by learning a uniform distribution over words, but such responses are obviously not useful. 
We evaluate the semantics of our responses in two ways: automated and through human evaluation.

\subsubsection{Automated evaluation}
For automated evaluation, we use the BLEU metric \cite{BLEU} on the generated responses of the test set. 
BLEU is commonly used for response evaluation on the MultiWOZ dataset as it has been shown to correspond reasonably well with human judgements on this task~\cite{DeltaBLEU}. 
We evaluate the BLEU score for both the responses generated if no ground truth is used as input, i.e., the GT style vector set to zero, and if it is actually used. 
The second score indicates how much the model relies on the ground truth. 

\begin{table*}[ht]
	\caption{Automatic evaluation results.}
	\label{tab:result_table_BLEU_diversity}
	\centering
	\resizebox{1.\textwidth}{!}{
	\begin{tabular}{lr@{ / }lcccc}
		\toprule
		& \multicolumn{2}{c}{} & \multicolumn{2}{c}{Diversity \textit{context}} & \multicolumn{2}{c}{Diversity \textit{stochastic beam search}}\\
		\cmidrule(r){4-5}\cmidrule(r){6-7}
		Experiment & \multicolumn{2}{c}{BLEU} & Distinct-2 & Distinct-1 & Distinct-2 & Distinct-1\\
		\midrule
		(1) Context + Slots & $29.97\%$ & -- & -- & -- & 0.170 & 0.098\\
		(2) Context + Slots + Template & $31.94\%$ & -- & -- & -- & 0.169 & 0.096\\
		(3) Context (proto) + Slots + Template & $31.43\%$ & -- & -- & -- & 0.169 & 0.094 \\
		(4) GT + Context (proto) + Slots + Template & $\mathbf{31.69}\%$ & $\mathbf{36.05}\%$ & 0.454 & 0.227 & 0.161 & 0.086\\
		(5) GT + Context + Slots + Template & $31.51\%$ & $33.08\%$ & 0.418 & 0.220 & 0.162 & 0.081\\
		(6) GT + Slots + Template & $31.56\%$ & $34.66\%$ & \textbf{0.485} & \textbf{0.237} & 0.165 & 0.086\\
		\bottomrule
	\end{tabular}}
\end{table*}

\subsubsection{Human evaluation}
\label{sec:experiments_human_evaluation}

We cannot totally rely on BLEU to evaluate the semantics of generated responses, even though BLEU is commonly used in both \ac{TDS} and other conversational modeling tasks~\cite{meng-2020-refnet,pei-2020-ecai}, because BLEU only evaluates the overlap between the generated responses and the demonstrated ground truth responses.
Therefore, we performed a human evaluation where a human is presented with a conversation and six generated responses for the last action. 
The responses had to be evaluated based on four metrics: \textit{Grammaticality}, \textit{Naturalness}, \textit{Context awareness} and \textit{Semantic correctness}. 
The first metric, \emph{Grammaticality}, aims to judge the English grammar and sentence structure. 
\emph{Naturalness} measures how ``human-like'' a response appears to be rather than automatically generated. 
This includes the usage of superfluous glue phrases like ``Sure'' or ``Certainly,'' and combining multiple chunks of information into more complex sentence structures. 
To have a natural conversation, it is also crucial to take the previous conversation and especially the user's question into consideration. 
Thus, the metric \emph{Context awareness} captures whether the generated responses fit into the conversation or not. 
Lastly, we also want to ensure that the semantics of the template response is left unchanged which is judged by the \emph{Semantic correctness}. 
We want that responses of different styles still communicate the same message. 
For this metric, we also provide the ground truth response from the human agent in the MultiWOZ dataset. 
Thus, semantic correctness can be considered as a human-evaluated BLEU metric.

A full overview of the human evaluation template and description can be found in the appendix.

\subsection{Implementation details}
\label{sec:experiments_hyperparams}

We use the Adam optimizer \cite{Adam} with a learning rate of 1e-4 for all of our models.
We use dropout \cite{Dropout} with a rate of 0.2 throughout the network to reduce overfitting. 
In addition, we start training with a teacher forcing ratio of 0.95, and reduce it exponentially such that it reaches 0.8 after 50k iterations (maximum training steps). 
The hidden size of the LSTMs is set to 512, as well as the response semantic size. 
For the context and response, we use four prototypes each and a size of 256 and 64, respectively. 
We sample $N=8$ times for every instance in the test dataset by alternating the prototype distribution of \OurModel{}. 
Specifically, we sample the attention distribution, which is used for creating the weighted sum over prototypes, by a Dirichlet prior with $\alpha=0.25$. 
Note that the template, slots and context are kept fixed for all 8 generated responses.
We keep the size of the ground-truth influenced style small in order to bias the network to also focus on the context. 
Experiments with more prototypes, e.g., six or eight, resulted in similar performance, but suffered more from overfitting and learned relations between possible ground truth prototypes and dialogue actions.

We want the ground truth to be considered as ``extra'' information and not necessary to generate a valid, grammatical response. 
We experienced that first words such as ``Sure'' and ``Okay'' mostly depend on the context, and should therefore be modeled in the context style vector. 
To ensure this, we use a two-step dropout strategy to augment the response paraphrasing noise vector during training. 
In 40\% of the cases, we set the response paraphrasing noise vector to 0. 
For the remaining 60\%, we sample from a geometric distribution with $p=0.4$ to determine until which generation time step we set the response paraphrasing noise to 0. 
This means that in $p=40\%$ of the cases we set the paraphrasing noise to 0 for the first 0 steps, i.e., we provide the ground-truth features during the whole generation process. 
In $(1-p)^{1}p=24\%$, we set it to zero only for generating the first token, and so on. 
This augmentation pushes the network to focus on the context style vector to generate the first few words.
The percentages were fine-tuned for the specific dataset, and might slightly differ for other conversation types, especially if the context plays less of a role. 


\section{Results and Analysis}

\subsection{Performance in terms of diversity}
\label{s05-1}

\begin{table*}[ht]
	\caption{Human evaluation results.}
	\label{tab:result_table_human_eval}
	\centering
	\resizebox{1.\textwidth}{!}{
	\begin{tabular}{lccccccccc}
		\toprule
		Metric & \multicolumn{3}{c}{\OurModel{} vs Stochastic Beam Search} & \multicolumn{3}{c}{\OurModel{} vs Human Responses} & \multicolumn{3}{c}{Stochastic Beam Search vs Human Responses}\\
		\cmidrule(l){2-4} \cmidrule(l){5-7}\cmidrule(l){8-10}
		&wins&ties&losses&wins&ties&losses&wins&ties&losses\\
		\midrule
		Grammaticality & \phantom{0}93 & 54 & 103 & \phantom{0}87 & 51 & 112 & \phantom{0}96 & 43 & 111\\
		Naturalness & 100 & 40 & 110 & \phantom{0}98 & 44 & 108 & \phantom{0}98 & 43 & 109\\
		Semantic correctness & 104 & 49 & \phantom{0}97 & \phantom{0}95 & 47 & 108 & \phantom{0}94 & 50 & 106\\
		Context awareness & 105 & 41 & 104 & 105 & 45 & 100 & 103 & 33 & 114\\
		\bottomrule
	\end{tabular}}
\end{table*}

To address RQ1, we conduct experiments by comparing different variants of \OurModel{} with a stochastic beam search.
The variants of \OurModel{} are different combinations of the following inputs:
\begin{enumerate}[label=(\arabic*),leftmargin=*,nosep]
\item Context: previous dialogue utterances.
\item Slots: slots that should be included in the final response.
\item Template: sampled response template that is used by \OurModel{} to extract style information.
\item GT: ground truth response, only used during training.
\item Context (proto): context with applied prototype layer.
\end{enumerate}
The results of all variants as well as the stochastic beam search baseline are listed in Table~\ref{tab:result_table_BLEU_diversity}.

First, in terms of diversity \OurModel{} outperforms the stochastic beam search baseline by a large margin.
Specifically, Distinct-2 is improved by around 0.3 while Distinct-1 is improved by around 0.15.
It has been shown that stochastic beam search significantly improves beam search \cite{DiversityHighQualitySeq2Seq}, and that it achieves around 1.5 times more distinct unigrams and up to 3 times more distinct bigrams per sentence compared to standard beam search. 
This means that stochastic beam search is already a strong method in terms of diversifying response generation.
\OurModel{} outperforms stochastic beam search by a large margin, which means that \OurModel{} generates more diverse responses than post-processing methods like stochastic beam search.
A major drawback we have experienced with stochastic beam search is that its diversity decreases over time/training iterations. 
This means that the longer we train, the lower the diversity of stochastic beam search. 
In contrast, for \OurModel{} diversity increases over time.

Second, the variants (4)--(6) achieve comparable performance in terms of diversity.
GT + Slots + Template achieves the best performance in terms of both Distinct-1 and Distinct-2.
However, when using the context prototypes as inputs, the diversity performance drops a bit.
The model needs to take into account the coherence with context through context prototypes by generating some context-aware words, which will hurt diversity a little bit.
E.g., for a context utterance starting with ``Can you \ldots'', the model will usually generate responses starting with ``Okay'' or ``Sure''.
Interestingly, we find that the performance drops a lot when using the original context instead of context prototypes.
When investigating the context style attention distributions on the context utterances, we see that the model focuses on names like ``Liverpool'' or specific times such as ``18:00''.
Using context prototypes solved this problem.
Furthermore, the training loss is significantly lower than that of the model with prototypes.
This indicates that the model overfits on specific contexts, and pays less attention to the ground truth style vector.
So the generated response will be less diverse.
For example, for the original context ``Is it possible that \ldots'', the response will mostly likely start with ``Yes''.
However, when using prototypes of this original context, e.g., ``Can you \ldots'', the response will not be limited to start with ``Yes'' any more.

\subsection{Performance in terms of semantics}
\label{s05-2}

Turning to RQ2, although \OurModel{} achieves significant improvements in terms of generating diverse responses, this does not necessarily imply that \OurModel{} can create a better user experience in practical systems.
In an extreme case, we can randomly select/generate responses to get near perfect diversity metrics, but the responses are useless because they lack semantic coherence, which cannot help users to achieve their task goals.
To this end, we also conduct experiments to evaluate the semantics of the generated responses.

On the one hand, we report automatic evaluation metric BLEU to check the overlap between generated responses and the demonstrated ground truth responses.
The results are listed in Table~\ref{tab:result_table_BLEU_diversity}.
We can see that \OurModel{} gets comparable results by using prototypes guided paraphrasing.
Specifically, variant (3), Context + Slots + Template, is the baseline here without using prototypes or incorporating paraphrasing noise.
By adding the context prototypes, we see that the BLEU score of variant (4) drops only 0.51\%, which is acceptable.
Also, by further adding the paraphrasing noise part, the diversity score of variant (5) increases a little bit compared to variant (4), with only 0.23\% lower than variant (3).
A possible reason is that it helps the model to do a better semantic modeling by teaching the model to separate semantic and style information.
To sum up, prototypes guided paraphrasing will not hurt semantics much in terms of BLEU scores. 

On the other hand, we also do human evaluation to further confirm the performance of \OurModel{} in practice.
Specifically, for each conversation, we present two responses generated by \OurModel{} (with the configuration: GT + Context (proto) + Slots + Template with sampled response  prototypes) and two by the stochastic beam search baseline. 
Furthermore, we also randomly select two corresponding human written responses, and replace their slots accordingly, which, we expect, will have high scores for Grammaticality, Naturalness, Semantic correctness, and Context awareness. 
Overall, we obtain scores for 6 responses for 250 conversation instances.
We compare the different models against each other by measuring the number of wins (i.e. higher metric score), ties and defeats over instances w.r.t. the four metrics.
The results are shown in Table~\ref{tab:result_table_human_eval}.
We have the following observations.

First, there are no significant differences for \OurModel{} and stochastic beam search, which means they achieve comparable performance in terms of the four metrics and can provide a satisfactory user experience in terms of the four evaluation aspects.
\OurModel{} is slightly worse than stochastic beam search in terms of Grammaticality and Naturalness.
The reason is that by incorporating prototype guided paraphrasing noise, it becomes harder for \OurModel{} to take care of the syntactic and grammatical issues of generated responses because there is a lot of noise in prototypes.
But \OurModel{} is comparable to stochastic beam search in terms of Context awareness and is slightly better than it in terms of Semantical correctness.
This means \OurModel{} can better guarantee the response semantics, which is consistent with the conclusion from Table~\ref{tab:result_table_BLEU_diversity}.
Note that \OurModel{} can provide much more diverse responses at the same time.

Second, although both \OurModel{} and stochastic beam search get satisfactory results, both perform worse than Human Responses.
This confirms the reliability and trustworthiness of the human evaluation results in Table~\ref{tab:result_table_human_eval}.
Note that for all the experiments in this paper, we assume that the correct system actions (slot and values to be included in the responses) are provided, which makes it easier for the model to generate the responses.
However, in practice, even the template-based systems will give improper or incomplete system actions sometimes, so we would expect even worse performance than Human Responses in real systems.
This means that there is still room for further improvement.
An exception is that \OurModel{} gets better performance than Human Responses in terms of Context awareness.
We believe the reason is that there are a number of cases where the human written responses seem to use/base certain templates in the MultiWOZ dataset, which makes them worse in terms of Context awareness.

\subsection{Visualization of style and semantic attention}
\label{s05-3}

\begin{figure}[h]
\centering
\includegraphics[width=.48\textwidth]{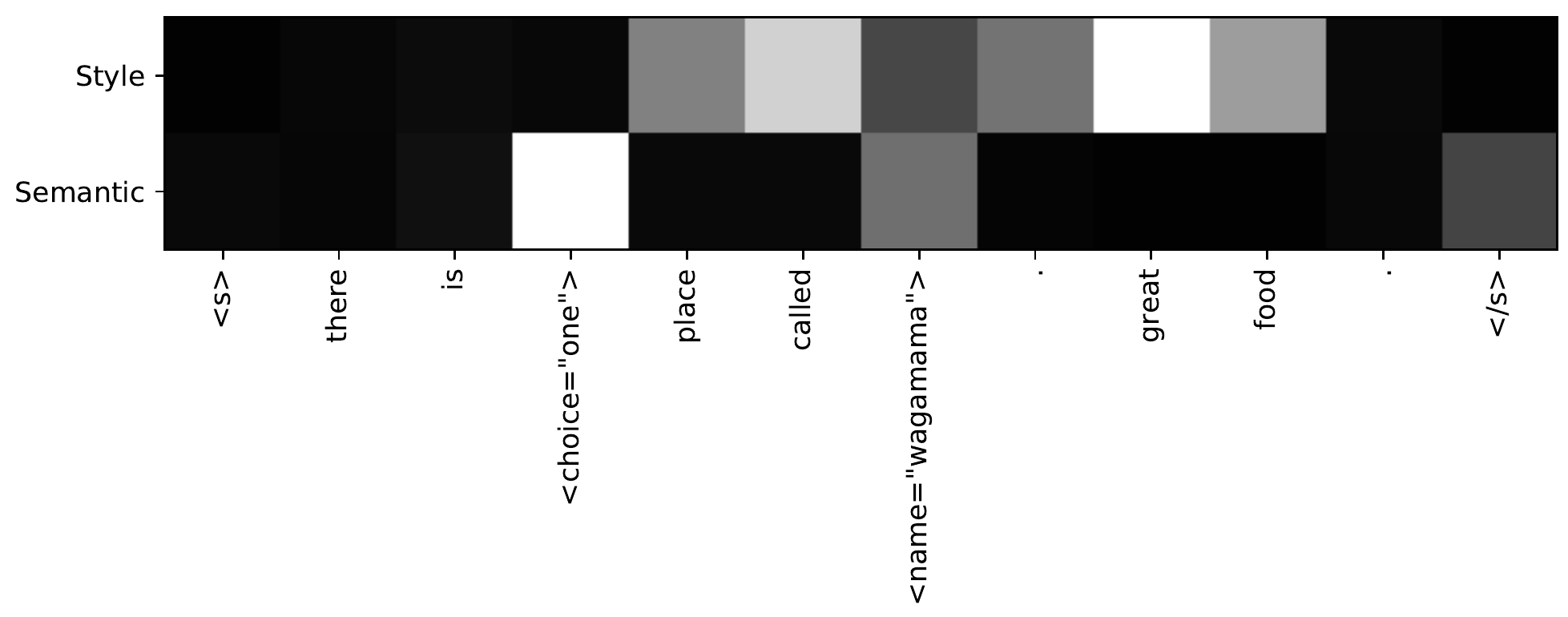}
\includegraphics[width=.48\textwidth]{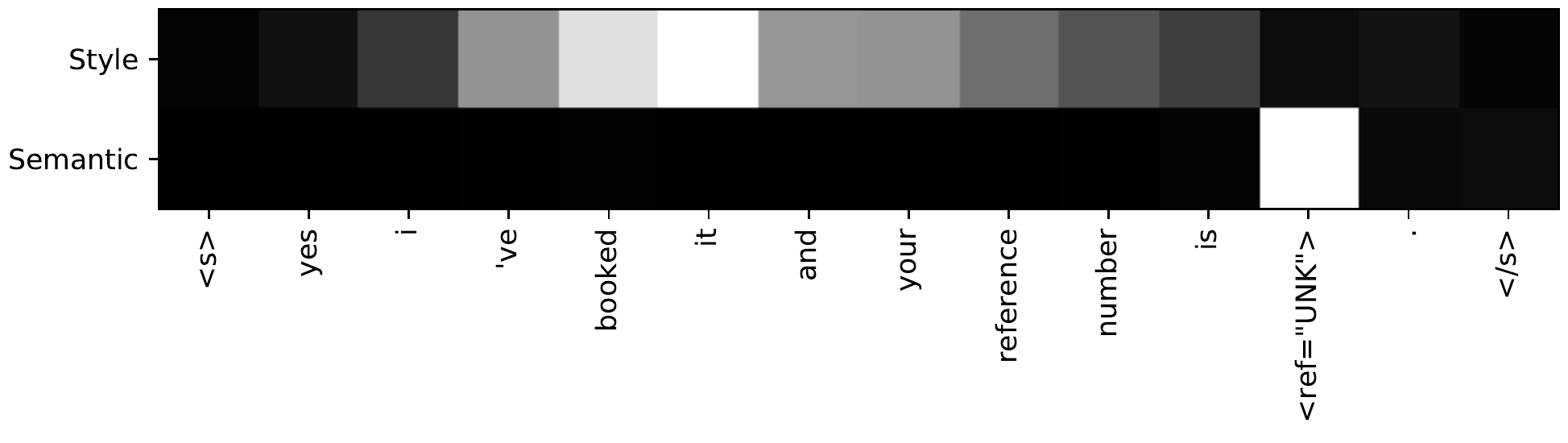}
\caption{Style and semantic attention visualization. Lighter color means higher attention weights.}
\label{fig:style_semantic_attention}
\end{figure}

To see whether \OurModel{} can correctly extract style and semantic information from prototypes and template, respectively, we visualize the style and semantic attention of two examples in Figure~\ref{fig:style_semantic_attention}.

\begin{table*}[h]
\caption{Qualitative analysis with case studies.}
\label{tab:case_study}
\centering

\resizebox{1.\textwidth}{!}{
\begin{tabular}{@{}c p{8.2cm} p{8.2cm}@{}}
\toprule
\multirow{13}{*}[-0.7cm]{\rotatebox[origin=c]{90}{Good case}} 
& \multicolumn{2}{l}{\textbf{Dialogue action:} Offer two choices for booking a table at a restaurant.}                                   \\ 
                            & \multicolumn{2}{l}{\textbf{Template:} I could try the \texttt{<name="charlie chan">}, or \texttt{<name="the golden house">} for you, if you wish .} \\
                            & \multicolumn{2}{l}{\textbf{Context:} Can you book a table for seven people on Thursday at 15:00 ?}                                     \\
                            & \multicolumn{2}{l}{\textbf{Slots:} \texttt{<name="La Mimosa">}, \texttt{<name="Shiraz">}}                                                            \\
                            \cmidrule(l){2-3}
                            & \multicolumn{1}{c}{\textbf{Diverse generations from \OurModel{}}}       & \multicolumn{1}{c}{\textbf{Diverse generations from stochastic beam search}}   \\
                            \cmidrule(l){2-2}\cmidrule(l){3-3}
                            & (1) Would you like to try La Mimosa or Shiraz?                 & (1) You can choose from La Mimosa , Shiraz .                                                                      \\
                            & (2) The La Mimosa and Shiraz are both available                & (2)  You can choose from La Mimosa or Shiraz.  \\
                            & (3) What about La Mimosa or Shiraz?                            & (3) you can choose from La Mimosa, or Shiraz. \\
                            & (4) I can recommend the La Mimosa or Shiraz.                   &  (4) Sure, we have the La Mimosa or Shiraz.                                                                     \\
                            & (5) Okay . Would you like to try La Mimosa or Shiraz?          & (5) You can choose from La Mimosa, or Shiraz.                                                                      \\
                            & (6) How about the La Mimosa or Shiraz?                         & (6) Sure, we have the La Mimosa and the Shiraz.                                                                      \\
                            & (7) I have the La Mimosa, Shiraz.                              & (7) Sure, we have the La Mimosa and the Shiraz. Do you want \\
                            & & \hspace{4mm}to book a of them?                                                                      \\
                            & (8) Okay, I have two options for you. La Mimosa and Shiraz.    & (8) Sure, we have La Mimosa and Shiraz. Do you want to go?                                                                   \\ \midrule
\multirow{14}{*}[-0.7cm]{\rotatebox[origin=c]{90}{Bad case}} 
 & \multicolumn{2}{l}{\textbf{Dialogue action:} Inform about trains leaving in a certain time frame}                                                                                          \\
                            & \multicolumn{2}{p{15.6cm}}{\textbf{Template:} Certainly, we have \texttt{<choice="many">} trains, the first train to arrive after \texttt{<arrive="17:36">} and the latest at \texttt{<arrive="18:45">}.}                                                                                              \\
                            & \multicolumn{2}{l}{\textbf{Context:} Yes, are there any trains leaving town after 13:45 on Friday?}                                                                                                \\
                            & \multicolumn{2}{l}{\textbf{Slots:} \texttt{<choice="several">}, \texttt{<arrive="16:07">}, \texttt{<arrive="24:07">}}                                                                                                    \\
                            \cmidrule(l){2-3}
                            & \multicolumn{1}{c}{\textbf{Diverse generations from \OurModel{}}}       & \multicolumn{1}{c}{\textbf{Diverse generations from stochastic beam search}}   \\
                            \cmidrule(l){2-2}\cmidrule(l){3-3}
                            & (1) I have several trains. One arrives at 16:07 and the other at 24:07. &  (1) There are several trains that fit your criteria. One arrives at 16:07 and the other at 24:07.                                                                    \\
                            &  (2) There are several trains, arriving by 16:07 or arriving at 24:07.                                                              &  (2) There are several trains that fit your criteria. One arrives at 16:07 and the latest at 24:07.                                                                   \\
                            &  (3) There are several trains that would get you there at 16:07, or would you like to take one at 24:07?&  (3) There are several trains that fit your criteria. One will get you there by 16:07 and 24:07.                                                                     \\
                            &  (4) I have several trains that arrive by 16:07 and 24:07.                                                              &   (4)  There are several trains that fit your criteria. One will get you there by 16:07 and the other arrives at 24:07.                                                                  \\ \bottomrule
\end{tabular}}
\end{table*}

We can see from Figure~\ref{fig:style_semantic_attention} that the context style attention apparently focuses on general sentence structure, which means that when learning to extract style information, \OurModel{} focuses on tokens that are about how to express the same meaning in different speaking styles. 
For example, in Figure~\ref{fig:style_semantic_attention}, more attention is paid to `place called' and `great food' in the first example, and `'ve booked it and your reference number' in the second example.
In many cases, context style attention is also related to the first words, which we hope to be captured by the context, e.g., beginnings like `yes' or `It' are often paid more attention.
When the response consists of two sentences, \OurModel{} often has attention on the first words of the second sentence and/or on the `</s>' token. 
This indicates \OurModel{} also encodes whether we use one or two sentences in a response.

We also visualize the semantic attention of the same examples when they are used as template responses.
The results are also shown in Figure~\ref{fig:style_semantic_attention}.
We can see that the semantic attention distribution is totally different as when the two examples are used as context prototypes.
In this case, \OurModel{} attends mostly on the slot values, which represent the semantics of the sentences.
In the first example, the number of places `\texttt{<choice="one">}' and the name of the place `\texttt{<name="wagamama">}' get the most attention.
And in the second example, the reference number of the booking `\texttt{<ref="UNK">}' gets the most attention.
Conversely, the other tokens are mostly ignored, this indicates that \OurModel{} indeed tries to extract semantics from template responses.

As a result, \OurModel{} can extract different styles from the prototypes and extract semantics from the template responses (which is output from the template-based \ac{TDS} systems), and combine them to get diverse responses while keeping the semantics unchanged.

\subsection{Qualitative analysis}
\label{s05-4}

To intuitively analyze the diverse responses from \OurModel{} and stochastic beam search, we list some examples in Table~\ref{tab:case_study}.

From the good case, we can see that although the responses from both models all look good in terms of semantics, the responses from \OurModel{} are much more diverse in terms of speaking styles.
For instance, both \OurModel{} and stochastic beam search will use different sentence patterns such as statements and questions, but \OurModel{} will generate different styles for statements, e.g., ``\ldots\ are both available'', ``I can recommend \ldots'', ``Okay, I have two options for you \ldots'', and for questions, e.g., ``Would you like to try \ldots?'', ``What about \ldots?'', ``How about \ldots?'', which are more diverse and human-like when applied to practical systems.
Conversely, the responses from stochastic beam search are less diverse.
Most responses are statements, and the speaking styles of those statements are not changed too much, e.g., ``You can choose from \ldots'' occurs 4 times and ``Sure, we have \ldots'' occurs 4 times as well.

There are also some bad cases for both models which need further improvements.
For the bad case in Table~\ref{tab:case_study}, we see that:
\begin{enumerate*}[label=(\arabic*)]
\item The generated responses are not always precise or consistent in terms of semantics, e.g., in response (1) of \OurModel{}, there are ``several'' trains in the first sentence, however, it generates ``One \ldots\ and the other \ldots'' in the second sentence.
This happens for all 4 responses from the stochastic beam search.
More efforts are needed in making the semantics consistent to slot values when generating responses.
\item The models do not take the template into account  as much as expected.
Specifically, in the template, it mentions there are ``many'' trains and then it gives the first and latest train.
However, the slot values ``\texttt{<arrive="16:07">}'' and ``\texttt{<arrive="24:07">}'' are independent from the sentence structures, which (therefore) have little influence on the sentence structure during response generation.
And when generating the responses, both models regard the two slot values as the only options, which clearly ignores some semantics in the template.
\end{enumerate*}

In addition, in both good and bad examples, we see that stochastic beam search puts the slots almost always at the same position. 
The start is often the same because the beams are biased towards selecting slots early. 
During generation, the non-slot words from beam search often have a probability of less than 10\% due to the large vocabulary.
In contrast, slots tend to have a probability close to 100\% because of the small set of slots, and the binary classifier $p_{gen}$ is close to $1$ or $0$. 
Thus, beams with having slots early in the output have a significantly higher probability, and dominate the generation process.
In contrast, sampling prototypes in \OurModel{} does not suffer from this issue as we are not comparing different outputs on probabilities, but just sampling input styles.


\section{Conclusion and Future Work}
In this work, we propose to combine the merits of template-based \acf{DRG} and corpus-based \acf{DRG} in \acfp{TDS} by presenting \OurModel{} based on prototype guided paraphrasing.
\OurModel{} can learn to extract style information from prototypes and extract semantics from template responses.
By combining both during generating, \OurModel{} can generate more diverse responses while preserving the semantics of template responses.
Automatic and human evaluations as well as a qualitative analysis demonstrate the effectiveness of \OurModel{} in terms of generating more diverse and human-like responses.

A limitation of \OurModel{} is that, in some cases, \OurModel{} will generate inconsistent content in the response and neglect some semantics in the template responses, which is not reflected by the slots.
As for future work, on the one hand, we hope to incorporate mechanisms to address those issues~\cite{wen2015semantically}.
On the other hand, we want to study how to apply \OurModel{} to other domains and languages with minimum effort in creating new datasets using transfer learning~\cite{yu2018modelling} or meta learning techniques~\cite{socher2013zero,ijcai2019-437}.

\section*{Code and data}
The dataset and the code of all the methods used for comparison in this paper are shared at:\\ \url{https://github.com/phlippe/P2_Net}.

\section*{Acknowledgements}
This research was partially supported by
Ahold Delhaize
and 
the Innovation Center for Artificial Intelligence (ICAI).
All content represents the opinion of the authors, which is not necessarily shared or endorsed by their respective employers and/or sponsors.

\bibliographystyle{IEEEtranN}
\bibliography{IEEEabrv,reference}


\appendix

The interface used for the human evaluation is shown in Figure~\ref{fig:appendix_mturk_interface}.
The following text has been provided to the MTurk workers to guide them during labeling:

\begin{figure}
	\centering
	\includegraphics[width=\linewidth]{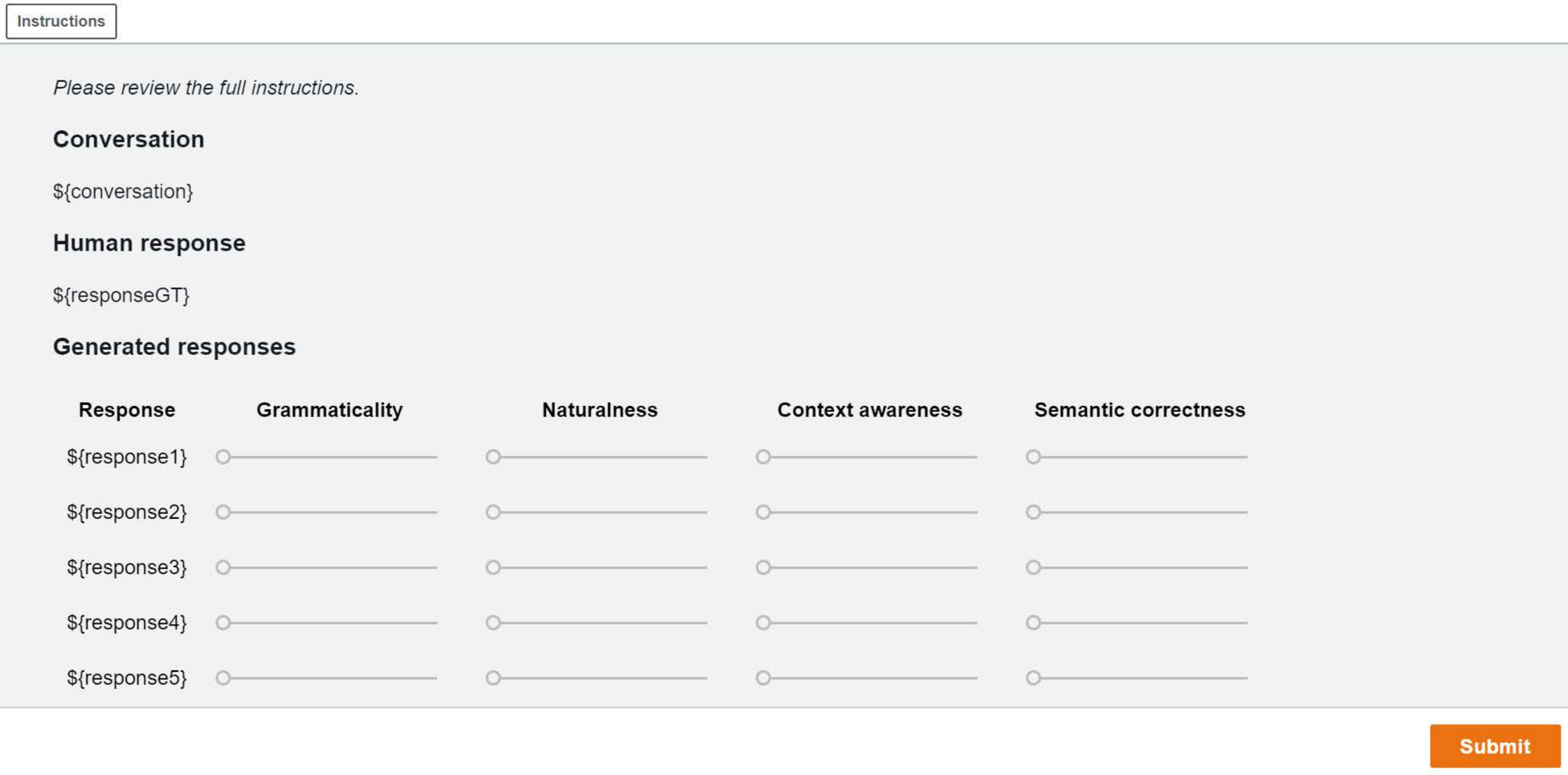}
	\caption{
		Interface for labeling during human evaluation. The variables of the form \$\{...\} are replaced by responses/context text depending on the specific instance to label.
	}
	\label{fig:appendix_mturk_interface}
\end{figure}

You are provided with a short conversation between a user and an agent, where the user asks for information, e.g., bookings of restaurants, hotels, attractions, taxis and trains. 
Both roles, user and agent, were performed by humans. 
The goal of this work is to replace the agent by an automated model which generates the responses instead. 
For evaluating the quality of the model, example responses need to be reviewed by humans, which is the aim of this MTurk project.
Below the conversation, you are provided with the next human response for the agent. 
This response can be seen as ``ground truth'', and all the generated responses should express the same content (more on review metrics below).
At the bottom of the page, you will find a table of 6 responses, each generated by a different model setting. 
Your task is to evaluate these responses based on four metrics: grammaticality, naturalness, context awareness, and semantical correctness.
All metrics are based on a scale of 1 (worst) to 3 (best) with a step size of 0.5. 
Below, the four metrics and the corresponding grading scheme for the values 1, 2 and 3 are explained in more detail. 
Please use the intermediate steps (1.5 and 2.5) if a response cannot be clearly assigned to a single bin.

\textbf{Grammaticality}: 
For this metric, you have to evaluate the response on the correct usage of the English language. 
This includes having a correct sentence structure, but also the correct tense and form of words. 
Note that this does not include capitalizing and lowercasing as for simplicity, most words were lowercased. 
Please use the following scale to evaluate the grammaticality of a sentence:
\begin{enumerate}[leftmargin=*,label=\arabic* - ]
\item The sentence is a (random) sequence of words without any clear structure, and is not understandable.\\
Example 1: ``The the the the the the''\\
Example 2: ``Drive hotel center I sorry''
\item The sentence contains a few minor mistakes such as repeating a word or wrong tense of a word. 
However, this does not significantly harm the understanding of the sentence. \\
Example 1: ``The the churchill college is in the west, and the address king's parade .''\\
Example 2: ``Yes, there is a museums, parks and boat tours. I would recommend a cinemas.''
\item The response is a valid sentence according to English language with max. 1 small (grammar) mistake.\\
Example 1: ``I would recommend the museum of classical art. It is located in the centre, address wollaston road.''\\
Example 2: ``I absolutely can! What type of entertainment are you looking for?''
\end{enumerate}

\textbf{Naturalness}: 
The ``naturalness'' of a sentence summarizes the characteristics of a response, which makes it sounds as a human would have said it, and not a typically machine-generated response. 
Thereby, we consider words that are not strictly necessary in a sentence, but make it sound more like a human conversation.
This includes words in the beginning like ``Sure'' and ``Certainly'', but also words like ``also'' or ``as well''.
Nevertheless, it is not required for a sentence to include any of these words to be considered as ``natural''. 
If the sentence/response could also have been written by a human, then the sentence is considered as natural. 
Note that natural sentences are not required to be free of any grammar or spelling mistakes as humans also make mistakes. 
Hence, ``naturalness'' can also be considered as ``humaneness'' of the response.
Please, use the following scale to evaluate the naturalness of a sentence:
\begin{enumerate}[leftmargin=*,label=\arabic* - ]
\item A sentence is considered as unnatural if it simply has a concatenation of short phrases and no words to connect those. 
It also includes unnatural orderings of information, as stating the detailed information before the general (i.e., phone number before the actual name).\\
Example 1: ``The phone number is 1234. The address is ABC street. The name is XYZ restaurant.''\\
Example 2: ``Center, ABC Street, Phone 1234, ''
\item The sentence is generic and is similar to a template response.\\
Example 1: ``I have found the Christian Art museum. The entrance fee is none.''\\
Example 2: ``The museum is located in the south. The phone number is 123000123.''
\item The sentence is possibly written by a human and/or cannot be distinguished from any other human response in terms of naturalness. 
It combines information into a single sentence and is less likely to be created from a template.\\
Example 1: ``The fitzwilliam museum is a free museum in the center of town, on trumpington street.''\\
Example 2: ``I absolutely can! What type of entertainment are you looking for?''
\end{enumerate}

\textbf{Context awareness}:
To have a natural, consistent conversation, the responses of the agent need to fit to the question of the user and the previous context of the conversation. 
Therefore, the metric ``context awareness'' measures whether a response takes the conversation so far into account or not. 
This includes both short-term dependencies, which directly relate to the last user's question, and long-term dependencies which model relations to earlier questions and/or responses. 
Please use the following scale to evaluate the context awareness of a sentence:
\begin{enumerate}[leftmargin=*,label=\arabic* - ]
\item The response is clearly not suitable for the user question, and assumes a different context.\\
Example 1: \\
User question: ``When is the last train leaving from Cambridge on Thursday?''\\
Generated response: ``Great choice! The last train is leaving at 5pm.''\\
Example 2:\\
User question: ``How much does the entrance costs to the cinema?''\\
Generated response: ``I absolutely can! It costs 4 pounds.''
\item The response is generic (i.e., can fit in almost every context) and unspecific to the question. \\
Example 1:\\
User question: ``Which cinema would you recommend?'' \\
Generated response: ``There are four cinemas in the south. The moonlight cinema has a 3.50 pounds entrance.''\\
Example 2:\\
User question: ``Can you tell me the phone number of the restaurant?''\\
Generated response: ``The phone number is 123000123.''
\item The response is specific to the question and fits well in the context. \\
Example 1:\\
User question: ``Which cinema would you recommend?''\\
Generated response: ``I can recommend the moonlight cinema, it only costs 3.50 pounds!''\\
Example 2:\\
User question: ``Ok, then I take the indian restaurant. Can you book me a table for four people?''\\
Generated response: ``Great choice! I reserved a table for you, the reference number is ABC123.''
\end{enumerate}

\textbf{Semantic correctness}:
For this metric, you need to compare the responses with the ``ground truth'' human response that is stated below the conversation. 
An optimal response only paraphrases the sentence, but expresses the exact same semantic/content. 
This includes the task (i.e., whether the agent tells the user the phone number, informs about a booking, asks for more specific information, etc.), the sentiment (i.e., ``Yes I can do it'' or ``No I'm sorry''), and mentioning all information (i.e., returns all information that was asked, like the phone number, address, etc.). 
Note that ungrammatical sentences that are hard or not understandable at all, should be evaluated with a low score as well as they do not reflect the same content and is not a paraphrase of the human response.
Please, use the following scale to evaluate the semantical correctness of a sentence:
\begin{enumerate}[leftmargin=*,label=\arabic* - ]
\item The response is mostly unrelated to the human response and/or does not answer the last question of the conversation. 
This also includes sentences which do not have a clear content and/or are not understandable. \\
Example 1:\\
Human response: ``Train ID123 leaves at 4pm.''\\
Generated response: ``I am sorry, there are no trains, there is ID123.''\\
Example 2:\\
Human response: ``I have the West Side hotel and the Cambridge suite. Which one would you like?''\\
Generated response: ``There are no hotels.''
\item The response is related to the human response, but differs in a few details such as missing out one information (e.g., phone number, postcode) or confusing numbers. \\
Example 1:\\
Human response: ``Train ID123 leaves at 4pm.''\\
Generated response: ``.''\\
Example 2:\\
Human response: ``I have the West Side hotel and the Cambridge suite. Which one would you like?''\\
Generated response: ``.''
\item The response is a paraphrase of the human response, representing the same content. \\
Example 1:\\
Human response: ``Train ID123 leaves at 4pm.''\\
Generated response: ``I have one train that leaves at 4pm, namely ID123.''\\
Example 2:\\
Human response: ``I have the West Side hotel and the Cambridge suite. Which one would you like?''\\
Generated response: ``Both the Cambridge suite and the West Side hotel are available.''
\end{enumerate}

\end{document}